\documentclass[journal]{IEEEtai}
\usepackage[colorlinks,urlcolor=black,linkcolor=black,citecolor=black]{hyperref}
\usepackage{amsmath,amsfonts}
\usepackage{algorithm}
\usepackage{array}
\usepackage{textcomp}
\usepackage{stfloats}
\usepackage{url}
\usepackage{verbatim}
\usepackage{graphicx}
\usepackage{cite}
\usepackage{graphicx}
\usepackage{subfigure}
\usepackage{multirow}
\usepackage{tcolorbox}
\usepackage{diagbox}
\usepackage{algorithmicx}

\usepackage{color}
\usepackage{amsthm,amsmath,amssymb}
\usepackage{mathrsfs}
\usepackage{hyperref}

\usepackage{algpseudocode}

\newcommand*{\revise}[1]{\textcolor{black}{#1}}

\usepackage {diagbox}

\usepackage{multirow} 
\makeatletter
\def\hlinewd#1{%
  \noalign{\ifnum0=`}\fi\hrule \@height #1 \futurelet
   \reserved@a\@xhline}

\algnewcommand{\algorithmicgoto}{\textbf{go to}}
\algnewcommand{\Label}{\State\unskip}

\begin{document}
%
\title{An Adaptive Federated Relevance Framework for Spatial Temporal Graph Learning}
%
%
%
%

\author{Tiehua Zhang$^{*}$,~\IEEEmembership{Member,~IEEE,}
        Yuze Liu$^{*}$,
        Zhishu Shen,~\IEEEmembership{Member,~IEEE,}
        Rui Xu,
        Xin Chen,
        Xiaowei Huang,
	Xi Zheng,~\IEEEmembership{Member,~IEEE,}
\thanks{Tiehua Zhang, Yuze Liu, Rui Xu, Xin Chen and Xiaowei Huang are with Ant Group, Shanghai, China (e-mail:\{zhangtiehua.zth, liuyuze.liuyuze, furui.xr, jinming.cx, wei.huangxw\}@antgroup.com).}
\thanks{Zhishu Shen is with the School of Computer Science and Artificial Intelligence, Wuhan University of Technology, Wuhan, China (e-mail: z\_shen@ieee.org).}
\thanks{Xi Zheng is with the Department of Computing, Macquarie University, Sydney, Australia (e-mail: james.zheng@mq.edu.au).}
\thanks{Corresponding author: Tiehua Zhang (tiehuaz@hotmail.com)}

\thanks{$^{*}$ co-first authorship.}
\thanks{Manuscript received XXX, 2022; }}

\markboth{IEEE Transactions on Artificial Intelligence, Vol. 00, No. 0, Month 2020}
{T. Zhang \MakeLowercase{\textit{et al.}}: IEEE Transactions on Artificial Intelligence}
%



\maketitle


\begin{abstract}
Spatial-temporal data contains rich information and has been widely studied in recent years due to the rapid development of relevant applications in many fields. For instance, medical institutions often use electrodes attached to different parts of a patient to analyse the electorencephal data rich with spatial and temporal features for health assessment and disease diagnosis. Existing research has mainly used deep learning techniques such as convolutional neural network (CNN) or recurrent neural network (RNN) to extract hidden spatial-temporal features. Yet, it is challenging to incorporate both inter-dependencies spatial information and dynamic temporal changes simultaneously. In reality, for a model that leverages these spatial-temporal features to fulfil complex prediction tasks, it often requires a colossal amount of training data in order to obtain satisfactory model performance. Considering the above-mentioned challenges, we propose an adaptive federated relevance framework, namely FedRel, for spatial-temporal graph learning in this paper. After transforming the raw spatial-temporal data into high-quality features, the core Dynamic Inter-Intra Graph (DIIG) module in the framework is able to use these features to generate the spatial-temporal graphs capable of capturing the hidden topological and long-term temporal correlation information in these graphs. To improve the model generalization ability and performance while preserving the local data privacy, we also design a relevance-driven federated learning module in our framework to leverage diverse data distributions from different participants with attentive aggregations of their models. In addition, we conduct extensive experiments on two real-world spatial-temporal datasets. The results demonstrate the effectiveness of our proposed framework in spatial-temporal interpretation, collaborative model training, and divergent data distribution handling in most settings of comparison.
\end{abstract}

\begin{IEEEImpStatement}
This research proposes a novel adaptive federated relevance framework, namely FedRel, for learning the spatial-temporal graph collaboratively without concerning the data privacy and scarcity of the spatial-temporal data (such as medical electorencephal data used at different institutions for disease diagnosis). FedRel takes the dispersion of non-IID data distribution from different participants into consideration, from which the relevance training algorithm is designed to drive the model updating for all participants. To better understand and incorporate the rich inter-dependencies topology and dynamic temporal changes in the spatial-temporal features of each participant, this research conceptualises the inter-intra graphs learning from the spatial-temporal features, which are generated through the designated feature transformation net. The dynamic inter-intra graph (DIIG) learning module is designed to capture the hidden topological and long-term temporal correlation information in these features. FedRel excels in both graph learning tasks and convergence speed in the collaborative matter.
\end{IEEEImpStatement}



\section{Introduction}
\IEEEPARstart{I}{n} recent years, spatial-temporal data has drawn increasing attention owing to its great potential in various fields and significant effects on daily activities such as traffic management~\cite{xu2020spatial}, health monitoring~\cite{zhang2017new}, and action recognition~\cite{wang2020traffic}. The advancement of deep learning has shed light on processing by extracting valuable features from spatial-temporal data to facilitate downstream prediction/regression tasks. For instance, traditional deep learning techniques such as convolutional neural networks (CNNs)~\cite{zhang2017new} and recurrent neural networks (RNNs)~\cite{shi2015convolutional, zheng2020hybrid} have been heavily investigated to uncover the hidden correlations and interdependencies in the spatial-temporal data sequence. 

Even though the existing models are able to capture and utilise the extracted correlation to some extent, many limitations start surfacing: CNN-based methods face challenges in discovering long-term temporal correlations, while RNN-based ones lack the capability to comprehend the global spatial structure. Additionally, one common issue is that both model types require grid-like data input, which is considered intractable when decoding the topological information from the spatial dimension. It is argued that the non-Euclidean graph data structure shows great potential in terms of representing inter-connectivity and dynamic trends in both spatial and temporal aspects~\cite{9530406}. Recently, modelling the spatial-temporal data into graph representation has risen to the spotlight owing to the development of graph neural network (GNN), reporting some encouraging improvements on both classification and regression tasks compared with CNN and RNN based models~\cite{jain2016structural,yu2017spatio}. It is also proven in prior work~\cite{wu2020connecting} that even for data in which explicit graph structure does not naturally exist, extracting and modelling the hidden graph structure from the original data could lead to a significant performance improvement. Following that, most existing methods~\cite{jain2016structural, yu2017spatio, seo2018structured, yan2018spatial} first transform spatial-temporal data into static graphs by modelling both nodes and edges, and then use GNNs to learn the embedding representations from the generated static graphs. Taking the traffic flow graph as an example in which different districts are treated as nodes, the flow of traffic between nodes will experience the changes at different hours of the day~\cite{xu2020spatial}. Ignoring dynamic correlations between nodes in the temporal dimension will undoubtedly lead to compromised and less desirable results. It is thus crucial to capture the inter-dependencies and dynamic temporal changes over time when turning the spatial-temporal data into the graph structure.

Apart from the quest to utilise spatial-temporal data more effectively, another pressing problem regards the data scarcity issue in real-world applications. For all learning-based tasks, insufficient data essentially causes an over-fitting problem of the model and makes it hard to generalise when unseen data appears during inference time. One common way to solve this problem is to collect a colossal amount of training data from multiple parties and train the model in a centralised manner. However, it has drawn increasing concerns over data privacy and security in many data-sensitive institutions. For instance, hospitals and drug research institutions rarely publish patient and clinical data due to their sensitivity, and trainable models thus fail to infer results that are not present in the training set. To this end, federated learning (FL)~\cite{kairouz2019advances} is introduced as a collaborative training scheme, which involves multiple parties without exposing local data to others. By integrating all participants' model weights or gradients, the model trained by FL will thus have a higher generalisation ability.


In the commonly used FL framework~\cite{mcmahan2017communication}, all participants' data is by default in the same distribution, while in real-world cases, the divergence of data distribution from different participants is nontrivial since it is determined by many uncontrollable factors like separate geographic locations. The most prevalent FL algorithm is called FedAvg~\cite{mcmahan2017communication}, in which the local model weights are sent to the server and aggregated in an iterative manner. However, it ignores the non-IID data distribution fact and assumes equal contribution from all participants. It is thus both intuitive and critical to incorporate the difference in data distribution from participants and make it part of the FL process to facilitate the training process. 

In this work, we propose a novel adaptive federated relevance framework called FedRel for spatial-temporal graph learning. This framework is composed of three key modules, including a feature transformation net to extract the important spatial-temporal features from raw data, a dynamic inter-intra graph (DIIG) module to generate the spatial-temporal graphs while capturing inter-dependencies and dynamic temporal changes across these graphs, and a relevance-based federated learning module to expedite the collaborative graph learning. It is worth mentioning that the intra-graph concept in our work refers to the spatial node correlations inside one generated spatial graph at any timestamp, while inter-graphs indicate temporal node correlations across multiple spatial graphs. The main advantages of our proposed framework are two folds: 1) DIIG uncovers the hidden topological graph structures of the transformed spatial-temporal features, which effectively encodes both spatial and temporal information into the embedding space when updating the node embedding through the GNN model; 2) The DIIGs from different participants are an integral part of the federated relevance training process. Thus the models are trained collaboratively to enable a better generalisation capability. 

The contributions of this work are summarised as follows: 
\begin{enumerate}

    
    \item[1.] We propose a novel adaptive federated relevance framework, namely FedRel, for spatial-temporal graph learning. The relevance of each participant is determined by adaptively attending the distance scores between local data distribution and the approximated global distributions, enabling the framework to put more attention on the participants with divergent data distributions and improve the model's generalisation ability.
    \item[2.] We implement a feature transformation net using a variety of model structures with different design philosophies, aiming to find the most practical model that excels in processing spatial-temporal raw data and is capable of generating high-quality spatial-temporal features.
    \item[3.] We conceptualise the inter-intra graphs learning from spatial-temporal features. We also design and implement a dynamic inter-intra graph (DIIG) module to generate the spatial-temporal graphs so as to capture inter-dependencies and dynamic temporal changes across these graphs.
    \item[4.] We conducted a comprehensive experiment of our framework on real-world dataset ISRUC\_S3~\cite{khalighi2016isruc} and SHL\_Small~\cite{wang2019enabling}, along with detailed ablation studies on each module for the interpretability purpose. The experimental results show that the proposed model achieves promising results in task-specific classifications, model convergence performance, and scalability. 
\end{enumerate}

\begin{figure*}[t!]
\centering
\includegraphics[width=\textwidth]{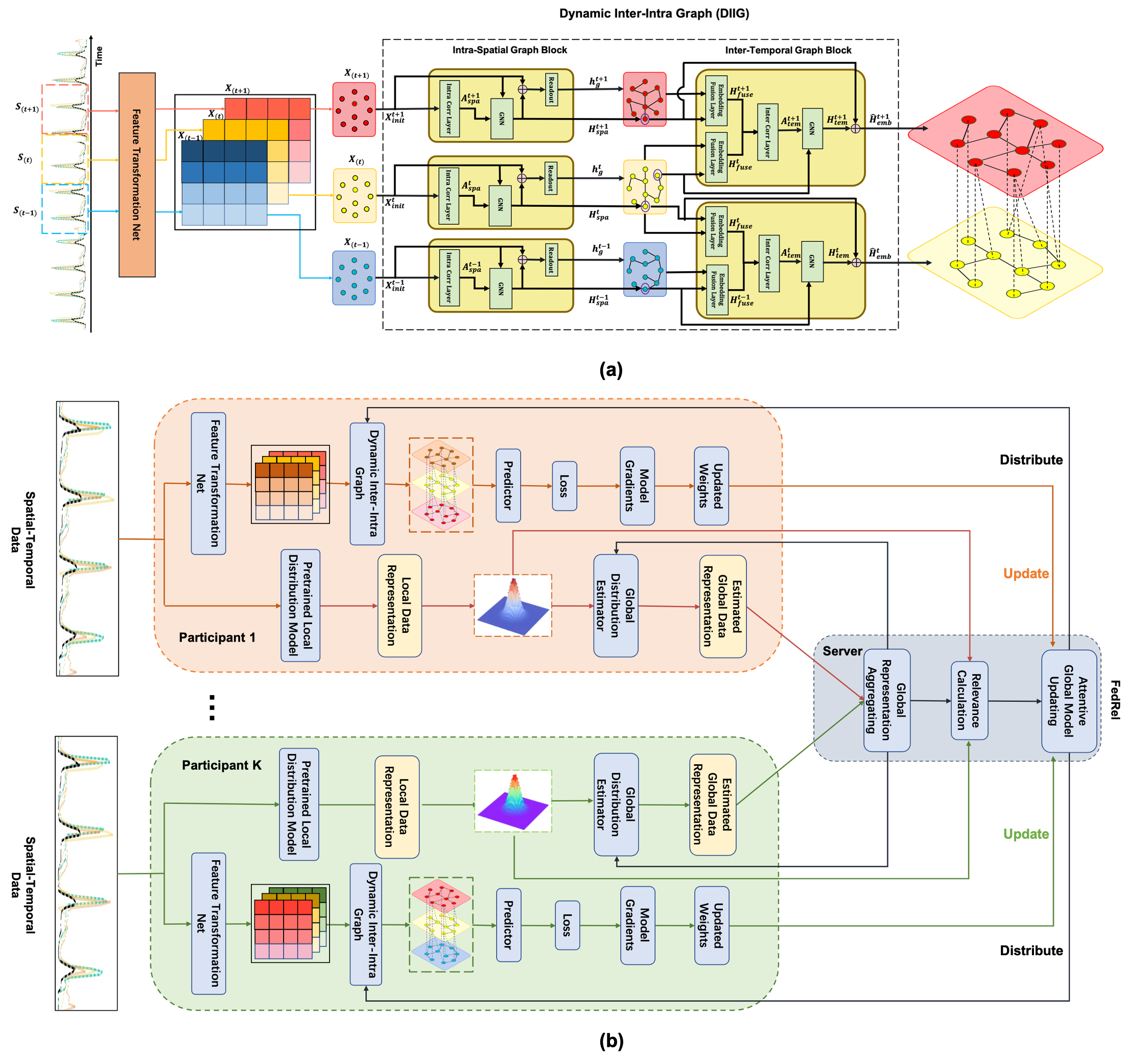} 
\caption{The overview of the FedRel (better viewed in color): (a) an example of exacting spatial-temporal features from the raw input data. In dynamic inter-intra graph (DIIG) module, The intra-spatial graph block first takes the initial node features at each timestamp to update embedding for each spatial graph. The inter-temporal graph block then merge across different spatial graphs in temporal dimension, producing the embedding with both spatial and temporal information; (b) the process of collaborative training on DIIGs from different participants. It enables an attentive model aggregation based on the relevance-guided divergence between local data distributions and approximated global data distribution.}
\label{fig:process}
\end{figure*}

\section{Related Work}

\subsection{Deep Learning for Spatial-Temporal Data}
The study of spatial-temporal data has attracted increasing attention in recent years, especially after the emergence of many advanced deep learning models. It is pointed out that spatial-temporal data could benefit applications in many disciplines such as emotion detection, traffic flow prediction, and sleep stage/quality classification~\cite{9530406}. However, It has been a long-standing challenge to better uncover and understand the hidden attributes in these numeric data. 

Recent research tries to interpret spatial and temporal dependencies, of which the deep learning model could take advantage to generate better prediction/classification results. For instance, a fast discriminative complex-valued convolutional neural network (CNN)~\cite{zhang2017new}, namely FDCCNN, is designed to apprehend the hidden correlations in the electroencephalogram (EEG) signals to expedite the sleep stage classification task. Alternatively, MLP-Mixer~\cite{huang2022mlp} is also used in the multi-channel temporal signal data, resulting in promising results on the regression task. It is proven effective to use recurrent neural network (RNN) based models for capturing non-linear interdependencies, including ConvLSTM~\cite{shi2015convolutional}, Bi-LSTM~\cite{zheng2020hybrid}. Following that, some studies start to use the attention mechanisms for capturing the long-term correlations~\cite{ienco2020deep}. Specifically, relevant spatial and temporal attention modules are designed to serve this purpose, reporting promising results in clustering multivariate time series data with varying lengths. {\color{black}Prior works~\cite{sun2019hierarchical,supratak2017deepsleepnet} also pointed out that time-invariant feature exists in spatial dimensions and showed the promising potential of designing separate feature modelling strategies for each dimension. Specifically, DeepSleepNet~\cite{supratak2017deepsleepnet} proposed a two-step algorithm that utilises CNN model to extract spatial features and Bi-directional Long Short-Term Memory (BiLSTM) to learn transition patterns. A hierarchical neural network is designed in~\cite{sun2019hierarchical} to realise spatial learning and time sequence learning as separate stages, respectively. Even though these research works presented significance and potential in terms of modeling spatial and temporal features, they are unable to model the underlying topological connectivity and dynamic time transitions simultaneously.}

\subsection{Graph Neural Networks (GNNs)}
Although applying CNN and RNN to spatial-temporal data has yielded promising results, the limitations are also clear. First, both of these techniques still require processing data in the Euclidean space, essentially ignoring the potential connectivity information in the spatial dimension. Also, the inter-dependent relations uncovered using both methods are difficult to interpret from human's perspective. The rise of 
GNN has shed light on this direction as some researchers have started investigating how to use graphs to form better topological representations in both spatial and temporal dimension~\cite{wu2020comprehensive}. 

However, constructing graph structure from the spatial-temporal data is the prerequisite before using GNN models, and there are two main hurdles when transforming the spatial-temporal sequence to GNN-required graph input: 1. to uncover the spatial correlations of the data streams to generate the adjacency matrix; 2. to extract the node features from the temporal values. A line of research has focused on solving this issue~\cite{jain2016structural, yu2017spatio, seo2018structured, yan2018spatial}. Specifically, graph convolutional recurrent network (GCRN)~\cite{seo2018structured} combines the LSTM network with ChebNet~\cite{defferrard2016convolutional} to handle spatial-temporal data. Structural-RNN~\cite{jain2016structural} uses node-level and edge-level RNN to uncover spatial correlations in the data. Alternatively, CNN could be used to embed temporal relationships to solve the exploding/vanishing gradient problems. For example, ST-GCN~\cite{yan2018spatial} uses the partition graph convolution layer to extract spatial information, while a one-dimension convolution layer is designed to extract temporal dependencies. Similarly, CGCN~\cite{yu2017spatio} combines a one-dimension convolution layer with a ChebNet or GCN layer, to handle spatial-temporal data. Regardless of using the CNN-, RNN-based models, or the recent advancements on GNN-based ones, the existing research is not able to utilise the underlying topological graph structures and temporal transition information in the spatial-temporal data simultaneously, and we intend to deal with this issue in our work.

\subsection{Federated Learning (FL) }
Federated Learning is a privacy-preserving collaborative learning paradigm designed to protect participants' local data privacy while enjoying a superior model performance. It enables the communication of participants' local gradients/weights to avoid exposing the raw data to others. FL is considered helpful in many fields, where data privacy is a major concern~\cite{zhang2021federated}. One of the widely used FL algorithms is FedAvg~\cite{mcmahan2017communication}, in which the local model weights are sent to the server and aggregated in an iterative manner. However, due to the diversity and divergence of local data distribution, the performance of FL is greatly compromised. It is pointed out in~\cite{zhao2018federated} that the non-IID data distribution from different participants leads to severe weight divergence throughout the training phase.
To cope with that, 
\cite{zhao2018federated, tuor2021overcoming, yoshida2019hybrid} propose to generate part of the global dataset by sharing some local data with the server so that the negative impact of non-IID could be mitigated. But the problem is obvious: it violates the design philosophy of FL by exposing partial data to others. Following that, Shin et al.~\cite{shin2020xor} proposes to upload XOR encoded seeds from participants' local data to the server for the global data approximation. The global data is decoded using these seeds to re-train the model on the server side. Instead of simply averaging the local model weights that cause the weight divergence, the contributions from different participants should be considered to facilitate the convergence of training. By calculating the distance between global and participant models at every communication round, attention scores are calculated as weighting coefficients on each local model when being aggregated~\cite{ji2019learning}. Instead of considering the weight distance or violating the local data privacy in other works, we believe the divergent local data distribution from different participants could contribute greatly when it comes to producing a well-performing model. In the FL process of our work, we intend to explore how the local data distribution and approximated global distribution could facilitate the training. The relevance of each participant is quantified and incorporated during training to generalise the model well.

Another line of research explores applying the GNN-based models to understand distributed spatial-temporal data in FL. To achieve an accurate traffic forecasting performance, Zhang et al.~\cite{ZhangIoT22} introduce GNN-based models that exploit the spatial correlations of the traffic graph in the FL-based system. Zhang et al.~\cite{ZhangTII21} follow this work in traffic prediction and propose an attention-based spatial-temporal GNN model under FL settings, which reports a good result in the traffic speed prediction task. Meng et al.~\cite{MengKDD21} propose a federated spatial-temporal model which explicitly encodes the underlying graph structure using GNN under the constraint of cross-node FL. This model can ensure the data generated locally remains decentralised without extra computation cost at the distributed edge devices. Compared with these, we propose to conceptualise the spatial-temporal correlation into an inter-intra graph learning module, from which both inter-dependencies among spatial dimensions and dynamic temporal changes can be better encoded into the embedding space.



\subsection{Discussion}
{\color{black}
The main issues in the existing work are two-fold:
\begin{enumerate}
\item The deep learning approaches like CNN-based (FDCCNN ~\cite{zhang2017new}) and RNN-based (ConvLSTM~\cite{shi2015convolutional}, Bi-LSTM~\cite{zheng2020hybrid}) are effective in processing spatial and temporal data separately. However, they have limitations when it comes to modelling the complex relationships and dependencies between different spatial and temporal entities. For example, it might be challenging for RNN-based approaches to model complex spatial relationships between different entities in a sequence, as they are designed primarily for modeling temporal dependencies. On the other hand, GNN-based methods (GCRN~\cite{seo2018structured}, ST-GCN~\cite{yan2018spatial}, CGCN~\cite{yu2017spatio}) enable the modeling of complex relationships between entities through the use of graphs. Nevertheless, they are not able to utilise the underlying topological graph structures and temporal transition information in the spatial-temporal data simultaneously, which impedes them from achieving better learning performance.
\item Federated learning algorithms like FedAvg~\cite{mcmahan2017communication}, FedP~\cite{ZhangIoT22} and FedAtt~\cite{ji2019learning} can realise collaborative training involving multiple parties without exposing local data to others to preserve data privacy. As an emerging field, the research that incorporates GNN-based approaches to federated learning is starting to gain attraction with several algorithms (FASTGNN~\cite{ZhangTII21} and CNFGNN~\cite{MengKDD21}) proposed recently. Still, it remains imperative to investigate methods for enhancing collaborative graph modeling capability on spatial-temporal attributes from the provided raw data~\cite{Liu2022FederatedGN}. 

\end{enumerate}
To solve the aforementioned problem, we introduce dynamic inter-intra graph learning to encode the spatial-temporal data. Specifically, intra-graph learning is designed to capture the spatial node correlations within a single graph snapshot generated at each timestamp, while inter-graph learning can learn the temporal node correlations across multiple graphs. Moreover, relevance-based federated learning is designed to expedite collaborative graph learning. The relevance of each participant herein is determined by adaptively attending the distance scores between local data distribution and the approximated global distributions. It thus puts more attention on the participants with divergent data distributions, from which the model’s generalisation ability can be improved.

}

\section{Preliminaries}

\begin{table}[t!]
\caption{Table of important notations}
\centering
\resizebox{\linewidth}{!}{
\begin{tabular}{c||c}
\hline
Notation & \ Description\\
\hline
\hline
$\oplus$ & Concatenation operation \\
$\left|\cdot\right|$ & The set size\\
$\left\|\cdot\right\|_{2}$ & L2 norm \\
$\emph{f}_{mn}(\cdot)$ & Row-wise mean function \\
$\mathcal{N}(\cdot)$ & Multivariate Gaussian function \\
$\sigma(\cdot)$ & Sigmoid activation function \\
$\emph{softmax}(\cdot)$ & Softmax function \\
$\exp(\cdot)$ & Exponential function \\
$\emph{MSE}(\cdot)$ & Mean square error function \\
$\emph{LN}(\cdot)$ & Layer normalisation \\
$\emph{msg}(\cdot)$ & Message passing function \\
$\emph{readout}(\cdot)$ & Graph readout function \\
$\emph{g}_{\theta}(\cdot)$ & Global distribution estimator \\
$\mathcal{L}$ & Loss function \\
w & Window size \\ 
N & The number of channels/nodes \\
K & The number of participants \\
T & Time steps \\
D & Raw data signal dimension \\
d & Feature dimension\\
L & Largest message passing layers \\
$\mathcal{S}$ & A set of raw spatial-temporal raw data \\
$\boldsymbol{S}$ & Spatial-temporal feature matrix \\
$\boldsymbol{s}_{n}$ & A row of raw signal frequency from $\boldsymbol{S}$ \\
$\mathcal{X}$ & A set of initial node features  \\
$\boldsymbol{X}$ & Initial node feature matrix of a graph \\
$\boldsymbol{x}_{i}$ & Initial feature vector of node $\emph{v}_{i}$ \\
$\mathcal{G}$ & A graph \\
$\mathcal{V}$ & The set of nodes in the graph \\
$\mathcal{E}$ & The set of edges in the graph \\
$\emph{v}$ & A node $\emph{v}\in\mathcal{V}$ \\
$\emph{e}_{i,j}$ & Edge between $\emph{v}_{i}$ and $\emph{v}_{j}$ \\ 
$\boldsymbol{A}$ & Adjacency matrix of a graph \\
$A_{\emph{i},\emph{j}}$ & Correlation value between nodes $\emph{v}_{i}$ and $\emph{v}_{j}$ \\
$\mathcal{G}_{spa}\left(t\right)$ & Spatial graph at $t$th time step \\
$\mathcal{G}_{tem}(t)$ & A stack of spatial graphs within a time window \\
$\Tilde{\boldsymbol{S}}^{(k)}$ & Reshaped local dataset at participants $k$\\
$\Tilde{\boldsymbol{s}}_{i}^{(k)}$ & A local data sample from $\Tilde{\boldsymbol{S}}^{(k)}$ at participant $k$\\

$\boldsymbol{H}$ & node embedding matrix at a graph \\
$\boldsymbol{h}_{i}$ & node $\emph{v}_{i}$'s embedding vector \\
$\Tilde{\boldsymbol{z}}$ & Predictor output \\
$\boldsymbol{y}$ & True label \\
$\boldsymbol{z}$ & Latent data representation of one local data point \\
$\boldsymbol{d}$ & Latent representation of local data \\
$\Tilde{\boldsymbol{d}}$ & Synthesised global data representation \\
$\boldsymbol{I}$ & Identity matrix \\
$\boldsymbol{W}$,$\Theta$,$\theta$,$\theta_{S}$ & Learnable model parameters\\
\hline
\end{tabular}
}
\label{tab:notation}
\end{table}

\subsection{Spatial-Temporal Graph}
A spatial-temporal data can be represented as 
$\mathcal{S} = \left\{\boldsymbol{S}\left(1\right),\boldsymbol{S}\left(2\right),...,\boldsymbol{S}\left(T\right)\right\}\in\mathbb{R}^{T\times N\times D}$, where  
$\boldsymbol{S}\left(t\right) = \left[\boldsymbol{s}_{1}\left(t\right),\boldsymbol{s}_{2}\left(t\right),...,\boldsymbol{s}_{N}\left(t\right)\right]\in\mathbb{R}^{N\times D}$, $t\in\left[1,2,...,T\right]$, is the time series, and $\boldsymbol{s}_{n}\left(t\right)\in\mathbb{R}^{D}$, $n\in\left[1,2,...,N\right]$, represents the original signal feature dimension $D$ at $n$th channel in temporal context. $N$ indicates the number of channels in the spatial dimension (e.g., devices and sensors). For each $\boldsymbol{S}\left(t\right)$, the features generated by the feature transformation net is $\boldsymbol{X}\left(t\right) = \left[\boldsymbol{x}_{1}\left(t\right),\boldsymbol{x}_{2}\left(t\right),...,\boldsymbol{x}_{N}\left(t\right)\right]\in\mathbb{R}^{N\times d}$, where $d$ is the size of the extracted features (As shown in the left part of Fig.~\ref{fig:process}.a).

To model the inter-dependencies over time steps, we define a time window $w$ to capture the historical sequence in temporal context, i.e., $\left\{\boldsymbol{X}\left(t-w\right),...,\boldsymbol{X}\left(t\right)\right\},w\in\left[0,1,...,t-1\right]$. We define a graph as $\mathcal{G} = \left(\mathcal{V},\mathcal{E},\boldsymbol{A}\right)$,  where $\mathcal{V} = \left\{\emph{v}_{1},...,\emph{v}_N\right\}$ is the set of nodes, and $\mathcal{E} = \left\{\left(\emph{v}_{i},\emph{v}_{j}\right)|v_{i},v_{j}\in\mathcal{V}\right\}$ denotes the set of edges in the graph, which can be quantified by an adjacency matrix $\boldsymbol{A}\in\mathbb{R}^{\left|\mathcal{V}\right|\times \left|\mathcal{V}\right|}$. The $A_{i,j}>0$ means there exists an edge $e_{i,j}$ between $v_i$ and $v_j$, and $A_{i,j} = 0$ otherwise.

Since there is no explicit graph structure in the extracted feature $\boldsymbol{X}\left(t\right)$, we use spatial channels to represent nodes in our problem setting, meaning $\left|\mathcal{V}\right|$ = $N$, which then is referred as a spatial graph $\mathcal{G}_{spa}$. The temporal graph, on the other hand, is composed of a stack of spatial graphs in time window $w$, which is denoted as Given $\mathcal{G}_{tem}\left(t\right) =  \left[\mathcal{G}_{spa}\left(t-w\right),\mathcal{G}_{spa}\left(t-w+1\right),...,\mathcal{G}_{spa}\left(t\right)\right]$. 








\subsection{Graph Neural Networks (GNNs)}
GNN works on the graph structure data and presents an effective way of learning both node and graph embeddings. It requires both initial node features and adjacency matrix as the input, and conducts the layer-based message passing to aggregate neighboring nodes information and update embeddings~\cite{xu2018powerful}. Given any $\mathcal{G} = \left(\mathcal{V},\mathcal{E},\boldsymbol{A}\right)$, the message passing function can be formalised as a function $\emph{msg}\left(\cdot\right)$ with trainable weights:





\begin{equation}
    \boldsymbol{H}^{l} = \emph{msg}\left(\boldsymbol{A},\boldsymbol{H}^{l-1}\right)
\end{equation}

\begin{equation}
    \emph{msg}\left(\boldsymbol{A},\boldsymbol{H}^{l-1}\right) = \sigma\left(\boldsymbol{W}_{agg}^l\cdot\left(\boldsymbol{H}^{l-1}\oplus\frac{1}{N}\cdot\boldsymbol{A}\cdot\boldsymbol{H}^{l-1}\right)\right)
\end{equation}
where $\emph{l} = 1,..,L$ denotes the message passing layer. $\boldsymbol{h}^{l}_{i}\in \boldsymbol{H}^{l}$ is updated embedding of node $v_i$, and $\boldsymbol{W}_{agg}^l$ is the trainable weight matrix at that layer. Note it exists $\boldsymbol{H}^{0} = \boldsymbol{X}$ at the initial stage, using the initial node features as the embedding information. $\oplus$ is the vector concatenation operation, and $\sigma\left(\cdot\right)$ is the activation function.

To derive the graph representation after the iterative updates of node embeddings, a graph readout function can be defined as $\emph{readout}\left(\cdot\right)$:


\begin{equation}
    \boldsymbol{h}^{\mathcal{G}}=\emph{readout}\left(\boldsymbol{H}^{L},\boldsymbol{X}\right)
\end{equation}

\begin{equation}
\emph{readout}\left(\boldsymbol{H}^{L}, \boldsymbol{X}\right)= \sigma\left(
    \boldsymbol{W}_{pool}\cdot\left(\frac{1}{N}\sum_{i=1}^{N}\boldsymbol{h}_{i}^{L}\oplus\boldsymbol{x}_{i}\right)\right)
\end{equation}
where $\boldsymbol{H}^{L}$ denotes the up-to-date node embeddings after $L$th message passing, and $\boldsymbol{h}^{\mathcal{G}}$ denotes the embedding representation of the graph derived from $\boldsymbol{H}^{L}$. We use a trainable weight $\boldsymbol{W}_{pool}$ as the linear transformation pooling, which can also be replaced by alternative operations such as min-pooling, max-pooling, or attentive-pooling~\cite{gat}.

\subsection{Federated Learning (FL)}
Federated learning is a collaborative training protocol to learn a well-generalised model without exposing the local participants' data to others. It enables the trainable weight aggregation on the server side. FL encompasses several rounds of training, in which each participant uploads the locally trained model to the server for model aggregation (e.g., FedAvg algorithm~\cite{mcmahan2017communication}). The updated model is then distributed back to participants. FL repeats this training protocol until the model converges. The objective function to minimise at the server side is:
\begin{equation}
    \min_{\Theta}F\left(\Theta\right):\mbox{where}\; F^{\left(k\right)}\left(\Theta\right) = \sum_{k=1}^{K}r^{(k)} F\left(\Theta\right)^{\left(k\right)}
\end{equation}
where $K$ is the number of participants. $r^{(k)}\geq 0$ is the weighting coefficient of participant $k$ for attentively aggregating the uploaded weights, where $\sum_{k=1}^{K}r^{(k)} = 1$. $F(\Theta)^{(k)}$ is the local objective function of participant $k$ where $\Theta$ is the trainable weight sets at that participant. Note that in FedAvg the $r^{(k)}$ is set to be $1$/$K$, meaning this algorithm considers equal contribution of each participant.

\section{Details of FedRel}
The overview of the architecture of the FedRel is shown in Fig.~\ref{fig:process}. We elaborate on the proposed framework in this section and formalise the four key components of the framework: 1) transforming the raw spatial-temporal data to high-quality initial node features; 2) intra-spatial graph block to uncover the attentive spatial correlations of different nodes and update node embedding per spatial graph; 3) inter-temporal graph block to incorporate the inter-dependencies into node embedding among neighbouring spatial graphs in temporal dimension; 4) FL relevance module for attentive weight aggregation based on divergent data distribution. The important notations appeared in this paper are summarised in Table~\ref{tab:notation}.

\subsection{Feature Transformation Net}
Since GNN requires both node features and adjacency matrix, the framework first takes the spatial-temporal sequence data as the raw input, followed by the designated feature transformation net to generate the initial node features matrix $\boldsymbol{X}\left(t\right)$ at each time step $t$ (Fig.~\ref{fig:process}.a). The feature transformation net can be implemented using variants of prevalent deep learning models, and the commonly used ones are CNN and RNN-based models~\cite{lou2021stfl,shi2015convolutional,zheng2020hybrid}. It is already clarified in Section III-A that the raw sequence data $\mathcal{S} = \left\{\boldsymbol{S}\left(1\right),\boldsymbol{S}\left(2\right),...,\boldsymbol{S}\left(T\right)\right\}\in\mathbb{R}^{T\times N\times D}$, the feature transformation process can be defined as:
\begin{equation}
    \left\{\boldsymbol{X}\left(1\right),...,\boldsymbol{X}\left(t\right)\right\} = \emph{transform}\left(\left\{\boldsymbol{S}\left(1\right),...,\boldsymbol{S}\left(t\right)\right\}; \theta_{S}\right)
\end{equation}
where $\theta_{S}$ is the trainable parameters in the feature transformation net, and $\mathcal{X} = \left\{\boldsymbol{X}\left(1\right),\boldsymbol{X}\left(2\right),...,\boldsymbol{X}\left(t\right)\right\}\in\mathbb{R}^{T\times N\times d}$ is the high-quality output, i.e. the generated initial node features. Note the feature transformation is performed as the local process at each participant, so it serves as the local pre-trained model and differs the $\theta_{S}$ in different participants.

\subsection{Dynamic Inter-Intra Graph (DIIG) Module} 
The dynamic inter-intra graph encompasses two primary goals: 1) to enable the intra-graph embedding update through the intra-spatial graph block, which uncovers the hidden correlations between spatial nodes and update node embedding in spatial dimension; 2) to further encode temporal changes of these spatial graphs and update the node embedding with these inter-dependencies in the time axis, in which the inter-node correlations between two adjacent spatial graphs are also quantified in the inter-temporal graph block. 
\subsubsection{Intra-Spatial Graph Block}
The intra-spatial graph block defines a spatial graph at any time $t$ based on the generated initial node features $\boldsymbol{X}\left(t\right)\in\mathbb{R}^{N\times d}$ from the feature transformation net. As explained in Section III-B, we deploy GNN for intra-graph learning, while updating the embedding of each node based on the correlations with spatial neighbour. We have reported using different static correlation functions in our preliminary work~\cite{lou2021stfl}, including PCC, PLV, and $K$-NN. Simply, these static correlation functions take a node feature matrix as the input, and use different ways of quantifying a numeric value based on a pair of node features (details of each explained in Section V-A). However, the shortcomings of using the static functions are obvious: 1) the connectivity between nodes will not change along with dynamic changes of node embedding; 2) it is unable to form the correlations between nodes residing in different graphs. To solve that, we design a dynamic intra correlation layer $\emph{c}\left(\cdot,\:\cdot\right)$ to generate the correlation of two nodes, which is calculated as:
\begin{equation}
    A_{spa}^{i,j}\!\left(t\right)\! =\! \emph{c}\!\left(\boldsymbol{x}_{i}\!\left(t\right)\!,\!\boldsymbol{x}_{j}\!\left(t\right)\!\right)\! = \! \frac{\exp\left(\left(\boldsymbol{x}_{i}\left(t\right)\right)^{T}\!\cdot\!\boldsymbol{W}_{spa}\!\cdot\!\boldsymbol{x}_{j}\left(t\right)\right)}{\sum_{n = 1}^{N}\!\exp\!\left(\left(\boldsymbol{x}_{i}\left(t\right)\right)^{T}\!\cdot\!\boldsymbol{W}_{spa}\!\cdot\!\boldsymbol{x}_{n}\left(t\right)\right)}
\end{equation}
where $\boldsymbol{x}_{i},\:\boldsymbol{x}_{j}\in\mathbb{R}^{1\times d}$. $\boldsymbol{W}_{spa}\in\mathbb{R}^{d\times d}$ is a learnable weight matrix and used to project the feature vectors into the correlation space, from which a correlation scalar between node $\emph{v}_i$ and $\emph{v}_j$ can be learnt dynamically. The softmax operation serves two purposes: 1) to ensure the non-negative correlation value; 2) to normalise the value between 0 and 1. The iterative calculation of each pair in the feature matrix $\boldsymbol{X}\left(t\right)$ leads to the full adjacency matrix $\boldsymbol{A}_{spa}\left(t\right)$ of this spatial graph $\mathcal{G}_{spa}\left(t\right)$.

As mentioned in Section III-B, the embedding update of $\mathcal{G}_{spa}\left(t\right)$ using GNN can be formalised as:
\begin{equation}
    \boldsymbol{H}^{l}_{spa}\left(t\right) = \emph{msg}\left( \boldsymbol{A}_{spa}\left(t\right),\boldsymbol{H}_{spa}^{l-1}\left(t\right)\right)
\end{equation}
where $\emph{l} \!=\! 1,..,L$ denotes the message passing layer, and $\boldsymbol{H}^{0}_{spa}\left(t\right) = \boldsymbol{X}\left(t\right)$. We have $\boldsymbol{H}_{spa}^{L}\left(t\right)= \left[\boldsymbol{h}^{spa}_{1}\left(t\right),\boldsymbol{h}^{spa}_{2}\left(t\right),...,\boldsymbol{h}^{spa}_{N}\left(t\right)\right]$ as the up-to-date node embeddings in $\mathcal{G}_{spa}\left(t\right)$, and the graph embedding of which can be calculated as:
\begin{equation}
    \boldsymbol{h}^{\mathcal{G}}\left(t\right) = \emph{readout}\left(\boldsymbol{H}^{L}_{spa}\left(t\right),\boldsymbol{X}\left(t\right)\right)
\end{equation}

Eq.8 and Eq.9 essentially provide the embeddings of nodes and graphs at each timestamp, which are then taken as the input to the inter-temporal graph block for embeddings update in the temporal dimension.
\subsubsection{Inter-Temporal Graph Block}
After obtaining the embeddings of nodes and graphs at the spatial graph level, it is critical to incorporate the inter-dependencies between different spatial graphs in the temporal dimension.

As indicated by Fig.~\ref{fig:process}a, the inter-temporal graph block takes the embeddings of nodes and graph as input of the embedding fusion layer. The process can be defined as:
\begin{equation}
    \boldsymbol{h}^{fuse}_{i}\left(t\right) = \emph{fusion}\left(\boldsymbol{h}^{spa}_{i}\left(t\right),\boldsymbol{h}^{\mathcal{G}}\left(t\right)\right)
\end{equation}
where $\boldsymbol{h}^{spa}_{i}\left(t\right)$ denotes the $\emph{v}_i$'s embedding at spatial graph $\mathcal{G}_{spa}\left(t\right)$, and $\boldsymbol{h}^{\mathcal{G}}\left(t\right)$ is the embedding of this graph. The detailed fusion operations can be expanded as:
\begin{equation}
   \emph{fusion}\!\left(\boldsymbol{h}^{spa}_{i}\!\!\left(t\right)\!,\boldsymbol{h}^{\mathcal{G}}\!\!\left(t\right)\!\right)\!\!=\!\!\emph{LN}\left(\sigma\!\left(\boldsymbol{W}_{fuse}\!\cdot\!\left(\boldsymbol{h}_{i}^{spa}\!\!\left(t\right)\oplus\boldsymbol{h}^{\mathcal{G}}\left(t\right)\!\right)\!\right)\!\right) 
\end{equation}
where $\oplus$ iteratively concatenates the graph embedding with node embeddings within that graph, making sure the updated node embedding contains not only the local neighborhood but also global graph-level structure information. $\emph{LN}\left(\cdot\right)$ is the layer normalisation operation used to improve the training speed and mitigate the overfitting~\cite{ba2016layer}. The matrix form of the generated fusion embedding of one graph is $\boldsymbol{H}_{fuse}\left(t\right)= \left[\boldsymbol{h}^{fuse}_{1}\left(t\right),\boldsymbol{h}^{fuse}_{2}\left(t\right),...,\boldsymbol{h}^{fuse}_{N}\left(t\right)\right]$.

Similar to the intra correlation layer, the inter correlation layer between nodes in two consecutive spatial graphs can be formalised as:
\begin{equation}
\begin{split}
    A_{tem}^{i,j}\left(t\right)&=\emph{c}\left(\boldsymbol{h}^{fuse}_{i}\left(t\right),\boldsymbol{h}^{fuse}_{i}\left(t-1\right)\right) \\  
    &=\frac{\exp\left(\left(\boldsymbol{h}^{fuse}_{i}\left(t\right)\right)^{T}\cdot\boldsymbol{W}_{tem}\cdot\boldsymbol{h}^{fuse}_{i}\left(t-1\right)\right)}{\sum_{n = 1}^{N}\exp\left(\left(\boldsymbol{h}^{fuse}_{i}\left(t\right)\right)^{T}\cdot\boldsymbol{W}_{tem}\cdot\boldsymbol{h}^{fuse}_{n}\left(t-1\right)\right)}
\end{split}
\end{equation}

It can be observed from Eq.12 that the inter correlation layer dynamically quantify the latent correlations using fused node embeddings in temporal dimension. $A_{tem}^{i,j}$ represents the normalised correlation value of any $\emph{v}_i$ and $\emph{v}_j$ node pairs at adjacent time steps, and $\boldsymbol{A}_{tem}$ records the values of learned temporal adjacency matrix.



Inter-temporal graph block employs a recursive approach to capture inter-dependencies across different time steps, which can be formulated as:
\begin{equation}
    \begin{split}
    \boldsymbol{H}\!_{tem}\!\left(\!t\!\right)\!\!= &\emph{msg}\!\bigl(...\\
    &\emph{msg}\!\left(\!\boldsymbol{A}\!\left(t-\!w\!+\!2\right)\!,\!\emph{msg}\!\left(\boldsymbol{A}_{tem}\!\left(t-\!w\!+1\!\right)\!,\!\boldsymbol{H}\!_{fuse}\!\left(t\!-w\!\right)\!\right)\!\right)\!\!\bigl)
    \end{split}
\end{equation}
where $\boldsymbol{H}_{tem}\left(t\right)$ represents the updated temporal node embeddings at time step $t$. The depth of recursion indicates whether a long-term temporal dependency is needed, which is controlled by an adjustable time window $w$. Note that $w=0$ means the node embeddings will not be updated by inter-graph message passing, and there exists $\boldsymbol{H}_{tem}\left(t\right)=\boldsymbol{H}_{fuse}\left(t\right)$.

\revise{To sum up, having an intra- and an inter-layer separately inside DIIG module is beneficial to uncover the underlying correlations between spatial features while encoding temporal dependency of spatial graphs simultaneously, leading to comprehensive representation learning on spatial-temporal data.} Therefore, the final node embeddings are expected to comprise both  topological spatial and inter-dependent temporal information. It can thus be derived as:
\begin{equation}
    \hat{\boldsymbol{H}}_{emb}\left(t\right) = \sigma\left(\boldsymbol{W}_{o}\cdot\left(\boldsymbol{H}_{spa}^{L}\left(t\right)\oplus\boldsymbol{H}_{tem}\left(t\right)\right)\right)
\end{equation}
where $\boldsymbol{W}_{o}$ is a linear output layer. $\hat{\boldsymbol{H}}_{emb}\left(t\right)$ is input for the row-wise mean function $\emph{f}_{mn}\left(\cdot\right)$ to generate the logits of this graph $\Tilde{\boldsymbol{z}}\left(t\right)$, which can be used in downstream classification tasks.
\begin{equation}
    \Tilde{\boldsymbol{z}}\left(t\right) = \emph{softmax}\left(\emph{f}_{mn}\left(\hat{\boldsymbol{H}}_{emb}\left(t\right)\right)\right)
\end{equation}

Specifically, we adopt the binary cross-entropy loss function to train the graph classification tasks in this work. 
\begin{equation}
    \begin{split}
        \mathcal{L}_{DIIG} = &\sum_{i}\bigl[\Tilde{z_{i}}\left(t\right)\cdot\log y_{i}\left(t\right) \\
        &+\left(1-\Tilde{z_{i}}\left(t\right)\right)\cdot\left(1-\log y_{i}\left(t\right)\right)\bigr]
    \end{split}
\end{equation}

\subsection{Federated Relevance Module}
We also design the federated relevance module, in which different participants can train a well-generalised model by considering the divergent local data distributions.

Given $K$ participants, the local data at participant $k$ is denoted as $\mathcal{S}^{\left(k\right)}\in\mathbb{R}^{T^{(k)}\times N\times D}$ (as explained in Section III-A), which can be reshaped to $\Tilde{\mathcal{S}}^{\left(k\right)} = \left\{\Tilde{\boldsymbol{s}}_{t}^{\left(k\right)}\right\}_{t=1}^{T^{(k)}}\in\mathbb{R}^{T^{(k)}\times ND}$, and $T^{(k)}$ is the number of data points at participant $k$. As pointed out in related research~\cite{kingma2013auto,zhao2017towards}, it is intractable to compute local data distribution $p_{data}\left(\Tilde{\boldsymbol{s}}^{\left(k\right)}\right)$ due to the undifferentiable marginal likelihood. To circumvent this problem, we intend to discover a latent space where encoded representations $\boldsymbol{z}^{\left(k\right)}$ can be learned to characterise the local data distribution $p_{data}\left(\Tilde{\boldsymbol{s}}^{\left(k\right)}\right)$ instead. Variational autoencoder (VAE) is considered as an ideal fit for this scenario as it uses a multivariate Gaussian to model the distribution of the latent space, which is defined as $q_{\phi}\left(\boldsymbol{z}^{\left(k\right)}_{i}\mid\Tilde{\boldsymbol{s}}_{i}^{\left(k\right)}\right) = \mathcal{N}\left(\boldsymbol{z}^{\left(k\right)}_{i};\mu^{\left(k\right)}_{i},\sigma^{2\left(k\right)}_{i}\boldsymbol{I}\right)$\cite{kingma2013auto}\cite{zhao2017towards}. The variational parameter $\phi$ in the probabilistic encoder can be learnt beforehand. The mean $\mu_i$ and s.d. $\sigma_i$ are outputs of the probabilistic encoder.

We collect the latent representations $\boldsymbol{z}_{i}^{\left(k\right)}$ of each data point $\Tilde{\boldsymbol{s}_i}^{\left(k\right)}$. The distribution representation $\boldsymbol{d}^{\left(k\right)}$ of $\mathcal{S}^{\left(k\right)}$ can be derived from:
\begin{equation}
    \boldsymbol{d}^{\left(k\right)} = \frac{1}{T^{(k)}}\sum_{i = 1}^{T^{(k)}}\boldsymbol{z}_{i}^{\left(k\right)}
\end{equation}

We define a learnable global distribution estimator $g_{\theta}^{(k)}\left(\cdot\right)$ at each participant to approximate the global data representation, the approximate global representation from participant $k$ is calculated as:

\begin{equation}
    \hat{\boldsymbol{d}}^{\left(k\right)}= g_{\theta}^{(k)}\left(\boldsymbol{d}^{\left(k\right)}\right) 
\end{equation}
where $\theta$ is parameterised by an MLP. The update of $\theta$ is an integral part of the federated learning process and will be explained in the following part. 

Each participant then uploads the latent local distribution vector, approximated global distribution vector and trainable weight sets $\Theta^{\left(k\right)}$ of DIIG to the server side for global model updates.
\begin{equation}
    \Tilde{\boldsymbol{d}} = \frac{1}{K}\sum_{k=1}^{K} \hat{\boldsymbol{d}}^{\left(k\right)}
\end{equation}
\begin{equation}
    r^{\left(k\right)} =\frac{\exp\left(\left\|\hat{\boldsymbol{d}}^{\left(k\right)}-\Tilde{\boldsymbol{d}}\right\|_{2}\right)}{\sum_{k=1}^{K}\exp\left(\left\|\hat{\boldsymbol{d}}^{\left(k\right)}-\Tilde{\boldsymbol{d}}\right\|_{2}\right)}
\end{equation}
\begin{equation}
    \Theta = \sum_{k=1}^{K}\left(r^{\left(k\right)}\cdot\Theta^{\left(k\right)}\right)
\end{equation}

The server receives approximated global distribution vectors from different participants and uses a vector aggregator to synthesise the true global data representation $\Tilde{\boldsymbol{d}}$ (Eq.19). Afterwards, the relevance score of each participant can be quantified using any distance-based measure like Euclidean distance in our work. $r^{(k)}$ is the relevance score of participant $k$, it measures how far its local distribution representation is away from the synthesised global data distribution $\Tilde{\boldsymbol{d}}$ at the server side (Eq.20). Therefore, a larger score indicates this local data deviates from the global one, which needs to assign more attention to calibrate. Eq.21 refers to the attentive weight aggregation to produce the global model. The advantage of this process has two folds: 1) it reflects how relevant each participant is from the perspective of the data distributions, which follows the less relevant, less obtained weight strategy; 2) the relevance score is adjusted adaptively at different communication rounds to lead to improve the generalisation ability of the global model, in which non-IID data distributions exist in different participants.

The server then delivers the global model $\Theta$ and synthesised global data distribution representation $\Tilde{\boldsymbol{d}}$ to each participant for a local model update. The next communication round leverages both information from the server. We propose a distribution-aware loss function, which combines the loss of the updated local model and the local loss of the approximated global distribution vectors simultaneously. 



\begin{equation}
\mathcal{L}^{\left(k\right)} = \mathcal{L}^{\left(k\right)}_{DIIG}+\emph{MSE}\left(\hat{\boldsymbol{d}}^{\left(k\right)},\Tilde{\boldsymbol{d}}\right)
\end{equation}

\revise{The distribution-aware loss consists of two parts. The first term $\mathcal{L}_{DIIG}$ serves as the cross-entropy loss in the supervised graph learning norm, which is used to minimise the distribution divergence between the predicted label and the ground truth label. The second term $\emph{MSE}\left(\cdot,\cdot\right)$, on the other hand, serves as the regulariser to reduce the distance between local distribution representation on the participant side and estimated global distribution representation on the server end, from which the relevance score can be calculated dynamically at each communication round to help guide the model optimisation.}

\revise{Following that, we use gradient descent to optimise the training of the proposed framework. The gradients w.r.t DIIG and global distribution estimator at participant $k$ can be derived from $\mathcal{L}^{\left(k\right)}$ via $\mathbf{g}_{d}^{(k)}=\frac{\partial \mathcal{L}^{(k)}}{\partial \Theta^{(k)}}$ and $\mathbf{g}_{e}^{(k)}=\frac{\partial \mathcal{L}^{(k)}}{\partial \theta^{(k)}}$, where $\Theta$ and $\theta$ are trainable weight sets of DIIG and global estimator function $g_{\theta}\left(\cdot\right)$, respectively.} Afterwards, $\theta$ and $\Theta$ can be updated at each participant:
\begin{equation}
\theta^{(k)} =\theta^{(k)} - \gamma\mathbf{g}^{(k)}_{e}
\end{equation}
\begin{equation}
\Theta^{(k)} =\Theta^{(k)} - \gamma\mathbf{g}^{(k)}_{d}
\end{equation}
where $\gamma$ is the learning rate. The full process of the federated relevance training algorithm is described in Algorithm~\ref{ferel}.

\begin{algorithm}[t]
\caption{Federated Relevance Training}\label{ferel}
\begin{algorithmic}[1]
\Require the set of participant $\mathcal{K}$, $\Tilde{\mathcal{S}}^{\left(k\right)}$ is the reshaped spatial-temporal data at paticipant $k$, learning rate $\gamma$.

\State Initialise $\Theta^{\left(k\right)}$ and $\theta^{\left(k\right)}$ at each participant
\State Compute local data latent representation $\boldsymbol{d}^{\left(k\right)}$ at each participant
\For{each round t = 1,2,...,n}
\For{each $k \in \mathcal{K}$ (in parallel)}
\State $\hat{\boldsymbol{d}}^{\left(k\right)}= g_{\theta}^{(k)}\left(\boldsymbol{d}^{\left(k\right)}\right)$ \Comment{approximate global data vector Eq.18}
\State Upload $\Theta^{\left(k\right)}$, $\theta^{\left(k\right)}$ and $\hat{\boldsymbol{d}}^{\left(k\right)}$ to the server 
\EndFor

\State $\Tilde{\boldsymbol{d}} = \frac{1}{K}\sum_{k=1}^{K} \hat{\boldsymbol{d}}^{\left(k\right)}$ \Comment{server synthesise global data representation}
\State Compute relevance scores $\boldsymbol{r}$=\{$r^{(1)}$, $r^{(2)}$, $...$, $r^{(K)}$\} by Eq.20
\State $\Theta = \sum_{k=1}^{K}r^{\left(k\right)}\cdot\Theta^{\left(k\right)}$ \Comment{attentive aggregation}
\State Server distributes $\Theta$ and $\Tilde{\boldsymbol{d}}$ to each participant

\For{each $k \in \mathcal{K}$ (in parallel)}
\State Update local model $\Theta^{\left(k\right)}$ = $\Theta$
\State $\mathcal{L}^{\left(k\right)} = \mathcal{L}^{\left(k\right)}_{DIIG}+\emph{MSE}\left(\hat{\boldsymbol{d}}^{\left(k\right)},\Tilde{\boldsymbol{d}}\right)$ \Comment{compute loss for next round}
\State Compute estimator's gradient $\mathbf{g}_{e}^{(k)}=\frac{\partial \mathcal{L}^{(k)}}{\partial \theta^{(k)}}$
\State Compute DIIG's gradient $\mathbf{g}_{d}^{(k)}=\frac{\partial \mathcal{L}^{(k)}}{\partial \Theta^{(k)}}$
\State $\theta^{(k)} =\theta^{(k)} - \gamma\mathbf{g}^{(k)}_{e}$ 
\State $\Theta^{(k)} =\Theta^{(k)} - \gamma\mathbf{g}^{(k)}_{d}$ \Comment{local model update}
\EndFor
\EndFor
\end{algorithmic}
\end{algorithm}

\section{Experiment}
We conduct extensive experiments on real-world datasets to evaluate our proposed framework from different perspectives. We aim to answer the following research questions: 
\begin{itemize}
    
    \item \textbf{RQ1:} How does our proposed FedRel perform compared with the baselines (centralized implementation and FedAvg) on the basis of the same feature transformation net and correlation module?
    \item \textbf{RQ2:} How does the DIIG contribute to the performance improvement compared with other static connectivity methods?
    \item \textbf{RQ3:} Which commonly used models perform well in terms of extracting the feature information from the spatial-temporal data?
    \item \textbf{RQ4:} What is the best GNN option when being incorporated in the DIIG?
    \item \textbf{RQ5:} How does the time window $w$ in DIIG affect the performance of the framework?
\end{itemize}

\begin{table}[t!]
\vspace{-1mm}
\caption{Shared parameter setup in DIIG}
\centering
\resizebox{0.8\linewidth}{!}{
\begin{tabular}{c||c}
\hline
Name & \ Setup\\
\hline
\hline
Readout func (GNN) & 2 linear layers, 1 output layer   \\
Layer depth (GNN) & 2 \\
Loss function & Cross entropy loss   \\
Optimiser & Adam~\cite{kingma2014adam}\\
Learning rate & 1.5e-3 \\
Dropout rate & 0.3 \\
Weight initialiser & Xavier~\cite{glorot2010understanding}\\
Communication round & 150\\
\hline
\end{tabular}
}
\label{tab:hyperpara}
\end{table}
\subsection{Experiment Setup}

\begin{table*}[]
\caption{Summary of comparison results under FedRel for ISRUC\_S3 and SHL\_Small w.r.t F1 score. The best results are marked in bold and underlined.}
\label{tab:ablation_table}
\resizebox{\linewidth}{!}{
\begin{tabular}{cll|ccccc||ccccc}
\hline\hline
\multicolumn{3}{c|}{\multirow{2}{*}{}} 
& \multicolumn{5}{c||}{\textbf{ISRUC\_S3}} 
& \multicolumn{5}{c}{\textbf{SHL\_Small}}                                                             \\ \cline{4-13} 
\multicolumn{3}{c|}{}  & \multicolumn{1}{c}{\textbf{CNN}}  & \multicolumn{1}{c}{\textbf{MLP-Mixer}}  & \multicolumn{1}{c}{\textbf{ConvLSTM}} & \multicolumn{1}{c}{\textbf{Bi-LSTM}} & \multicolumn{1}{c||}{\textbf{STTransformer}} & \multicolumn{1}{c}{\textbf{CNN}}      & \multicolumn{1}{c}{\textbf{MLP-Mixer}} & \multicolumn{1}{c}{\textbf{ConvLSTM}} & \multicolumn{1}{c}{\textbf{Bi-LSTM}} & \multicolumn{1}{c}{\textbf{STTransformer}} \\\hline
\multicolumn{1}{c|}{\multirow{6}{*}{\textbf{\begin{tabular}[c]{@{}l@{}}GCN\end{tabular}}}} & 
\multicolumn{1}{l|}{\multirow{3}{*}{\textbf{\begin{tabular}[c]{@{}l@{}}Intra Only\end{tabular}}}}                & \textbf{$K$-NN}        &0.627                                      &0.702                                        &0.504                                        &0.591                                   &0.376                                           &0.659                                       &0.943                                       &0.960                                        &0.627                                   &0.596                                          \\
\multicolumn{1}{l|}{}  & \multicolumn{1}{l|}{}                                                                                & \textbf{PCC}         &0.775                                       &0.713                                        &0.632                                       &0.574                                   &0.496                                           &0.881                                       &0.947                                       &0.958                                        &0.621                                   &0.591                                          \\
\multicolumn{1}{l|}{}    & \multicolumn{1}{l|}{}                        & \textbf{PLV}         &0.723                                       &0.701                                        &0.579                                        &0.579                                   &0.487                                           &0.881                                       &0.958                                       &0.947                                        &0.631                                   &0.604                                          \\\cline{2-13}
\multicolumn{1}{l|}{}          & \multicolumn{1}{l|}{\multirow{3}{*}{\textbf{\begin{tabular}[c]{@{}l@{}}Inter-Intra\end{tabular}}}}                                        & \textbf{DIIG}$\left(w=2\right)$ &0.804                                       &\underline{\textbf{0.794}}                                        &\underline{\textbf{0.703}}                                        &\underline{\textbf{0.679}}                                      &0.612                                           &0.891                                       &\underline{\textbf{0.972}}                                       &\underline{\textbf{0.967}}                                        &\underline{\textbf{0.804}}                                      &0.625                                          \\
\multicolumn{1}{l|}{}                  & \multicolumn{1}{l|}{}                                         & \textbf{DIIG}$\left(w=3\right)$&\underline{\textbf{0.814}}                                       &0.792                                        &0.681                                        &0.655                                   &\underline{\textbf{0.618}}                                           &\underline{\textbf{0.893}}                                       &0.967                                       &0.957                                        &0.788                                   &\underline{\textbf{0.638}}                                          \\
\multicolumn{1}{l|}{}                   & \multicolumn{1}{l|}{}                                         & \textbf{DIIG}$\left(w=4\right)$&0.785                                       &0.768                                        &0.625                                        &0.634                                   &0.609                                           &0.887                                       &0.952                                       &0.952                                        &0.773                                   &0.609                                          \\\hline\hline
\multicolumn{1}{c|}{\multirow{6}{*}{\textbf{\begin{tabular}[c]{@{}l@{}}GAT\end{tabular}}}} & \multicolumn{1}{l|}{\multirow{3}{*}{\textbf{\begin{tabular}[c]{@{}l@{}}Intra Only\end{tabular}}}}    & \textbf{$K$-NN}        &0.812                                       &0.770                                        &0.585                                        &0.571                                   &0.402                                           &0.902                                       &0.957                                       &0.967                                        &0.625                                   &0.599                                          \\
\multicolumn{1}{l|}{}   & \multicolumn{1}{l|}{}                                             & \textbf{PCC}         &0.806                                       &0.818                                        &0.552                                        &0.591                                   &0.426                                           &0.909                                       &0.962                                      &0.952                                        &0.631                                   &0.620                                          \\
\multicolumn{1}{l|}{}    &\multicolumn{1}{l|}{}                                       & \textbf{PLV}         &0.816                                       &0.817                                        &0.563                                        &0.628                                   &0.420                                           &0.911                                       &0.962                                       &0.964                                        &0.651                                   &0.623                                          \\\cline{2-13}
\multicolumn{1}{l|}{}    &\multicolumn{1}{l|}{\multirow{3}{*}{\textbf{\begin{tabular}[c]{@{}l@{}}Inter-Intra\end{tabular}}}}                           & \textbf{DIIG}$\left(w=2\right)$ &0.837                                       &0.830                                       &0.682                                        &0.718                                   &0.630                                           &\underline{\textbf{0.931}}                                       &\underline{\textbf{0.981}}                                       &\underline{\textbf{0.974}}                                        &\underline{\textbf{0.894}}                                   &\underline{\textbf{0.635}}                                          \\
\multicolumn{1}{l|}{}  &\multicolumn{1}{l|}{}                                         & \textbf{DIIG}$\left(w=3\right)$&\underline{\textbf{0.855}}                                       &\underline{\textbf{0.831}}                                        &\underline{\textbf{0.712}}                                       &\underline{\textbf{0.719}}                                   &\underline{\textbf{0.628}}                                           &0.928                                       &0.973                                       &0.966                                        &0.888                                   &0.614                                          \\
\multicolumn{1}{l|}{}              & \multicolumn{1}{l|}{}                                        & \textbf{DIIG}$\left(w=4\right)$&0.827                                       &0.825                                        &0.693                                        &0.704                                   &0.615                                           &0.921                                       &0.958                                       &0.957                                        &0.886                                   &0.621                                          \\\hline\hline
\multicolumn{1}{c|}{\multirow{6}{*}{\textbf{\begin{tabular}[c]{@{}l@{}}GPS\end{tabular}}}}   &
\multicolumn{1}{l|}{\multirow{3}{*}{\textbf{\begin{tabular}[c]{@{}l@{}}Intra Only\end{tabular}}}}    &                                                              \textbf{$K$-NN}        &0.764                                      &0.679                                       &0.579                                      &0.541                                  &0.426                                       &0.866                                       &0.950                                       &0.958                                        &0.880                                   &0.596                                          \\
\multicolumn{1}{l|}{}      & \multicolumn{1}{l|}{}                                                                           & \textbf{PCC}         &0.731                                       &0.743                                        &0.565                                        &0.592                                   &0.388                                           &0.914                                       &0.946                                       &0.957                                        &0.892                                   &0.612                                          \\
\multicolumn{1}{l|}{}           & \multicolumn{1}{l|}{}                                                                      & \textbf{PLV}         &0.804                                       &0.710                                        &0.622                                        &0.547                                   &0.424                                           &0.902                                       &0.952                                       &0.961                                        &0.886                                   &0.590                                          \\\cline{2-13}
\multicolumn{1}{l|}{}  & \multicolumn{1}{l|}{\multirow{3}{*}{\textbf{\begin{tabular}[c]{@{}l@{}}Inter-Intra\end{tabular}}}}                              & \textbf{DIIG}$\left(w=2\right)$ &\underline{\textbf{0.808}}                                       &\underline{\textbf{0.748}}                                        &0.632                                        &\underline{\textbf{0.685}}                                      &\underline{\textbf{0.602}}                                          &\underline{\textbf{0.918}}                                       &\underline{\textbf{0.957}}                                       &\underline{\textbf{0.969}}                                        &\underline{\textbf{0.902}}                                      &\underline{\textbf{0.761}}                                          \\
\multicolumn{1}{l|}{}  &\multicolumn{1}{l|}{}                                          & \textbf{DIIG}$\left(w=3\right)$&0.792                                       &0.742                                        &0.641                                        &0.624                                   &0.552                                           &0.893                                      &0.948                                       &0.952                                        &0.895                                   &0.747                                          \\
\multicolumn{1}{l|}{}   & \multicolumn{1}{l|}{}                                                 & \textbf{DIIG}$\left(w=4\right)$&0.787                                       &0.768                                        &\underline{\textbf{0.647}}                                        &0.675                                   &0.583                                           &0.902                                       &0.945                                       &0.952                                        &0.891                                   &0.742                                          \\\hline\hline
\multicolumn{1}{r|}{\multirow{6}{*}{\textbf{\begin{tabular}[c]{@{}l@{}}GraphSAGE\end{tabular}}}}   &
\multicolumn{1}{l|}{\multirow{3}{*}{\textbf{\begin{tabular}[c]{@{}l@{}}Intra Only\end{tabular}}}}                                 & \textbf{$K$-NN}        &0.714                                       &0.721                                        &0.545                                        &0.570                                   &0.433                                           &0.887                                       &0.958                                       &0.961                                        &0.628                                   &0.642                                          \\
\multicolumn{1}{l|}{}     & \multicolumn{1}{l|}{}                                                                            & \textbf{PCC}         &0.741                                       &0.720                                        &0.564                                        &0.555                                   &0.391                                           &0.790                                       &0.956                                       &0.948                                        &0.628                                   &0.614                                          \\
\multicolumn{1}{l|}{}      &\multicolumn{1}{l|}{}                                                                           &\textbf{PLV}         &0.793                                       &0.729                                        &0.563                                        &0.581                                   &0.453                                           &0.887                                       &0.958                                       &0.954                                        &0.625                                   &0.579                                          \\\cline{2-13}
\multicolumn{1}{l|}{}  & \multicolumn{1}{l|}{\multirow{3}{*}{\textbf{\begin{tabular}[c]{@{}l@{}}Inter-Intra\end{tabular}}}}                             & \textbf{DIIG}$\left(w=2\right)$ &\underline{\textbf{0.829}}                                       &0.821                                        &\underline{\textbf{0.727}}                                        &\underline{\textbf{0.701}}                                   &0.596                                          &\underline{\textbf{0.922}}                                       &\underline{\textbf{0.968}}                                       &0.958                                        &\underline{\textbf{0.683}}                                   &\underline{\textbf{0.669}}                                          \\
\multicolumn{1}{l|}{}     & \multicolumn{1}{l|}{}                                         & \textbf{DIIG}$\left(w=3\right)$&0.820                                       &0.826                                        &0.715                                        &0.664                                   &\underline{\textbf{0.618}}                                              &0.901                                       &0.957                                       &\underline{\textbf{0.961}}                                        &0.639                                   &0.595                                         \\
\multicolumn{1}{l|}{}       & \multicolumn{1}{l|}{}                                     & \textbf{DIIG}$\left(w=4\right)$&0.798                                       &\underline{\textbf{0.828}}                                        &0.683                                        &0.696                                   &0.603                                        &0.914                                       &0.955                                       &0.962                                        &0.639                                   &0.613                                          \\\hline\hline

\end{tabular}
}
\end{table*}

\subsubsection{Dataset}
In our experiments, ISRUC\_S3~\cite{khalighi2016isruc} and SHL~\cite{wang2019enabling} are used as the benchmark dataset. ISRUC\_S3\footnote{\url{https://sleeptight.isr.uc.pt/?page_id=48}} collects polysomnography (PSG) recordings in 10 channels from 10 healthy subjects (i.e., sleep experiment participants). These PSG recordings are labeled with five different sleep stages according to the American Academy of Sleep Medicine (AASM) standard~\cite{jia2020graphsleepnet}, including Wake, N1, N2, N3 and REM. The SHL dataset contains multi-modal signal data from a body-worn camera and from 4 smartphones being equipped at various body locations. The SHL\_Small\footnote{\url{https://www.dropbox.com/s/gqlqugj6rpojq64/SHL-HUAWEI_10000s.npz.zip?dl=0}} is a concise version of SHL, which contains labelled locomotion signals with different motions, including Stay Still, Walk, Run, and Train.

\subsubsection{Data Partition}
In the experiment, we randomly select 80\% as the global training data and 20\% as the global test set. For the training data of each participant, we emulate the partial non-IID data setup in the real-world environment and draw the respective training data for each participant from this 80\% data pool~\cite{zhang2020achieving}. To verify the scalability of the proposed framework, we set the number of participants to be \{2, 3, 5, 10, 50\}, respectively. Note that the global test set is used to verify the model performance at each communication round. 

\subsubsection{Parameter \& FL Settings}
The shared parameter setup across different modules of the framework can be found in Table~\ref{tab:hyperpara}. Specifically, we use Adam~\cite{kingma2014adam} as the optimiser with the learning rate of 1.5e-3 for all trainable models in the framework. The dropout rate~\cite{srivastava2014dropout} is set to be 0.3. We apply the same readout function for all GNN models with the architecture of a shallow MLP with [32, 64, 64] neurons at each layer. Note the input node embedding size is 32, and the output graph embedding size is 64. Models are collaboratively trained for 150 communication rounds and the batch size is set to be 8 in both training and test data. All trainable parameters are initialised through Xavier~\cite{glorot2010understanding}.

\revise{To simulate the experimental setting of federated learning, we utilise the multiprocessing package\footnote{\url{https://docs.python.org/3.8/library/multiprocessing.html}} in Python, in which each participant is assigned a process in the experiment, including the server process as well. By doing so, each participant could realise the upload of local model weights, latent local distribution vector and approximated global distribution to the server in their own process over communication rounds. The attentive aggregation on the server side will synchronise local information uploads of participants at each communication round.}

\subsubsection{Comparison Methods}
As our framework is composed of three key components, we thus examine on using different models in the feature transformation part, spatial-temporal graph generation part, and federated learning part.  


We first consider different candidates in the feature transformation net, aiming to find the best match that is capable of deriving high-quality features from the raw spatial-temporal data, which could be used as the initial node embedding for the spatial-temporal graph generation part. We choose CNN~\cite{lou2021stfl} as the default feature transformation net as it has already proven effective in our preliminary work~\cite{lou2021stfl}. However, we re-implement other models, which are used to process the spatial-temporal related data in other works, to verify their capability of extracting features under our framework:

\begin{itemize}
    \item \revise{MLP-Mixer model is first introduced in~\cite{mlpmixer} for image classification tasks. Unlike traditional convolutional neural networks (CNNs), the MLP-Mixer model uses a series of linear mixer layers, each consisting of a multi-layer perceptron (MLP) with global average pooling, to process the spatial and channel dimensions of the input image separately. This allows the model to capture both local and global features of the image in a computationally efficient manner. It also shows promising results when applied to extract the spatial temporal features in~\cite{huang2022mlp}.
    }
    
    \item \revise{ConvLSTM~\cite{shi2015convolutional} is a recurrent neural network (RNN) based architecture that extends the traditional LSTM (Long Short-Term Memory) network by incorporating convolutional layers into the model. The input and hidden states of the LSTM in ConvLSTM are replaced with 3D tensors, where the third dimension corresponds to the channels (spatial) of the input. This allows the model to operate on input sequences with spatial information (e.g. images or videos) while capturing temporal dependencies in the data. ConvLSTM reports an excellent and consistent performance in capturing spatial-temporal correlations.
    }

    \item \revise{Bi-LSTM~\cite{zheng2020hybrid} also has a RNN-based architecture. It consists of two LSTM layers, one of which processes the input sequence in a forward direction and another that processes it in a backward direction. This allows the model to take into account both past and future contexts when making predictions. The RNN backbone in Bi-LSTM also helps to uncover the correlations in the spatial-temporal data.
    }
    \item \revise{Transformer~\cite{vaswani2017attention} is a type of sequence-to-sequence (Seq2Seq) model that uses self-attention mechanisms to process input sequences in parallel. It consists of two main parts: the encoder and the decoder. The encoder takes the input sequence, encodes it into a sequence of hidden states, and generates a context vector that summarizes the input sequence. The decoder then takes the context vector as input, and generates an output sequence one token at a time. When processing the spatial-temporal data, it calculates the attention of channels (spatial) at each time step to capture both spatial and temporal dependencies.
    }
    
\end{itemize}

We also explore different strategies to quantify the node correlations, which serve as the adjacency matrix of the spatial-temporal graph. As explained in Section IV-B, DIIG enables a trainable correlation uncovering and embedding update simultaneously, and other three static methods are $K$-Nearest Neighbor ($K$-NN), Pearson Correlation Coefficient (PCC) and Phase Locking Value (PLV). The different node correlation functions are described below:
\revise{
\begin{itemize}
    \item $K$-NN~\cite{jiang2013graph} generates the adjacency matrix that only selects the $k$-th nearest neighbor of each node to represent the node correlation of the graph. The values in the adjacency matrix are either 0 or 1.
    \item PCC~\cite{pearson1903laws} is known as the Pearson correlation function, which measures the similarity between each pair of nodes using the node features. It scales the values in the adjacency matrix between 1 and -1. 
    \item PLV~\cite{aydore2013note} is a useful tool to measure the phase relationship between two node signals (features). It returns a value between 0 and 1 to qualify the consistency of each pair of nodes in the adjacency matrix.
\end{itemize}
}
Regarding the GNN used in the DIIG for each participant (shown in Fig.~\ref{fig:process}.a), we compare the following models:
\revise{
\begin{itemize}
    \item GCN~\cite{gcn} is the first-order approximation to the spectral GNN model. It introduces a graph convolutional layer to aggregate messages from neighbor nodes.
    \item GraphSAGE~\cite{sage} is the first type of GNN that introduces the sampling technique. Each node only selects a subset of connected neighbor nodes when operating the message aggregation. This model reports a good performance on large-scale graph learning.
    \item GAT~\cite{gat} is the first work that incorporates the attention mechanism in GNN, computing attention scores for all connected node pairs for weighted message aggregation from neighboring nodes.
    \item GPS~\cite{gps} introduces an adaptive sampling technique, in which only important neighbor nodes are selected to participate in the message aggregation process. The level of importance of each node is tuned at each training iteration. This method reports a good performance in terms of graph representation learning and generalisation on unseen nodes.    
\end{itemize}
}
When it comes to the FL setup, we compare our proposed FedRel with different baselines listed below. Note that only the trainable weights in DIIG unit and the global distribution estimator at the participant side contribute to the federated learning process (as indicated in Fig.~\ref{fig:process}b), and the feature transformation net only serves as the pre-trained model, which shares the same parameters across different participants.
\revise{
\begin{itemize}
\item Centralized indicates the training of DIIG without FL settings, where data are stored in one place while being accessible to the model. 
\item FedAvg~\cite{mcmahan2017communication} is the first federated learning algorithm, which enables collaborative training of a number of local models from different participants. A central server receives uploaded model weights at each communication round and performs weight averaging. 
\item FedP~\cite{zhang2020achieving} only selects a fraction of participants (0.6 of full participants as default) to upload model weights at each communication round. It also uses FedAvg for the model update.
\item FedAtt~\cite{ji2019learning} proposes a layer-based attention calculation to quantify the contribution of each participant. It then uses attentive weight aggregating to update the global model at each communication round.
\end{itemize}
}



\begin{figure}[t]
\centering
\includegraphics[width=1.0\linewidth]{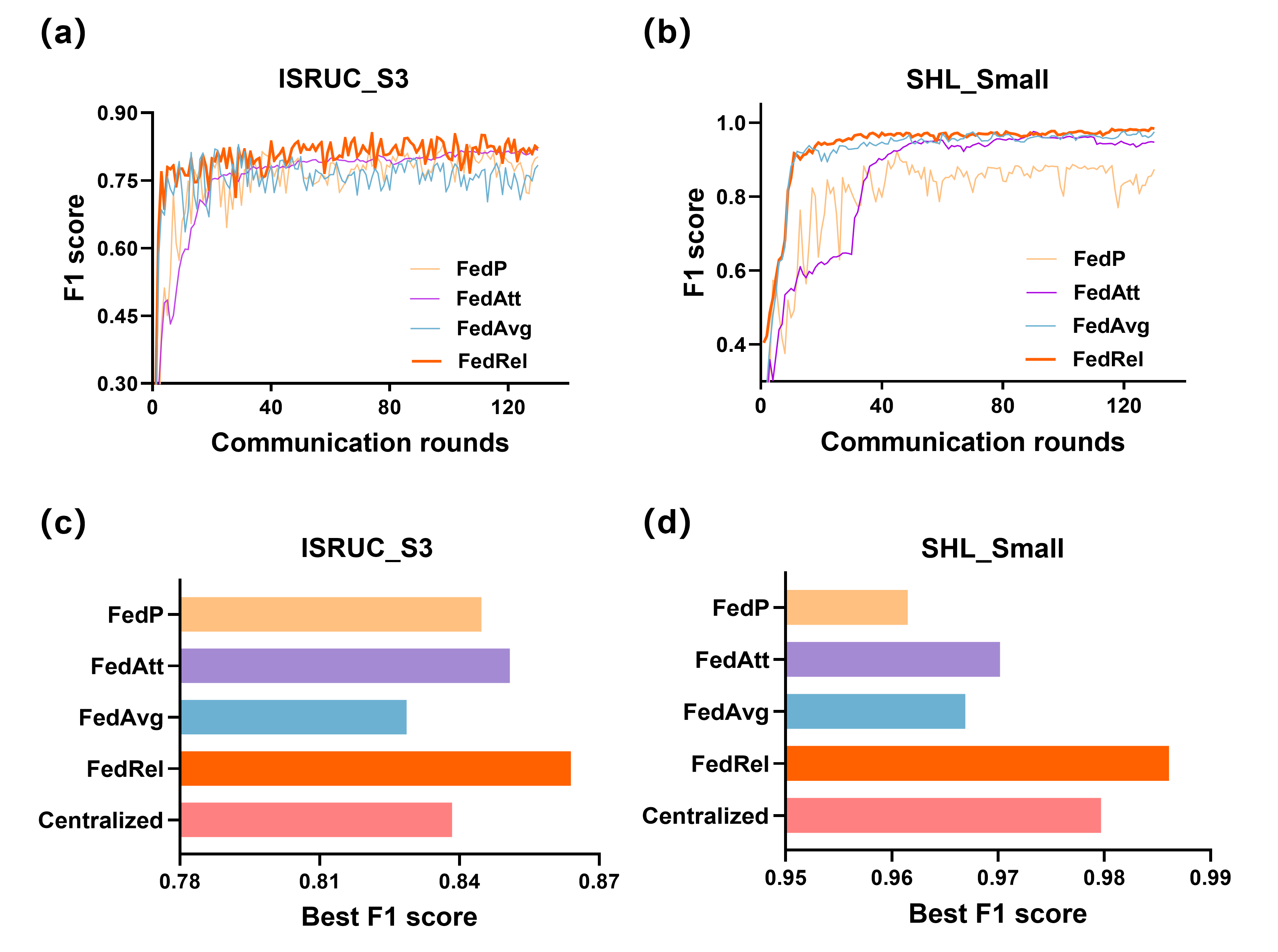} 
\caption{Comparison of the F1 of our framework with baselines FL models.}

\label{fig:FLResult}
\end{figure}

\subsection{Experiment Results and Analysis}
In this part, we evaluate the performance of our proposed framework by conducting comprehensive experiments and analysis. It essentially helps us to understand the significance of each module under different settings separately.
\subsubsection{Performance Analysis}
The performance of FedRel and DIIG are analysed separately.\newline
\textbf{FedRel Performance (RQ1):} We first compare the overall performance of our proposed framework with several baseline FL models, as well as the centralized setting. Note that all other modules, including the feature transformation net and GNN model, are cherry-picked based on the best results reported in Table~\ref{tab:ablation_table} (detailed analysis unveiled in ablation study). Herein, the framework is composed of CNN\_GAT\_DIIG\_FedRel ($w=3$) for ISRUC\_S3 and MLP-Mixer\_GAT\_DIIG\_FedRel ($w=2$) for SHL\_Small, both using GAT in DIIG module. For a fair comparison, other baselines are under the same module setup.

\begin{figure}[t]
\centering
\includegraphics[width=0.85\linewidth]{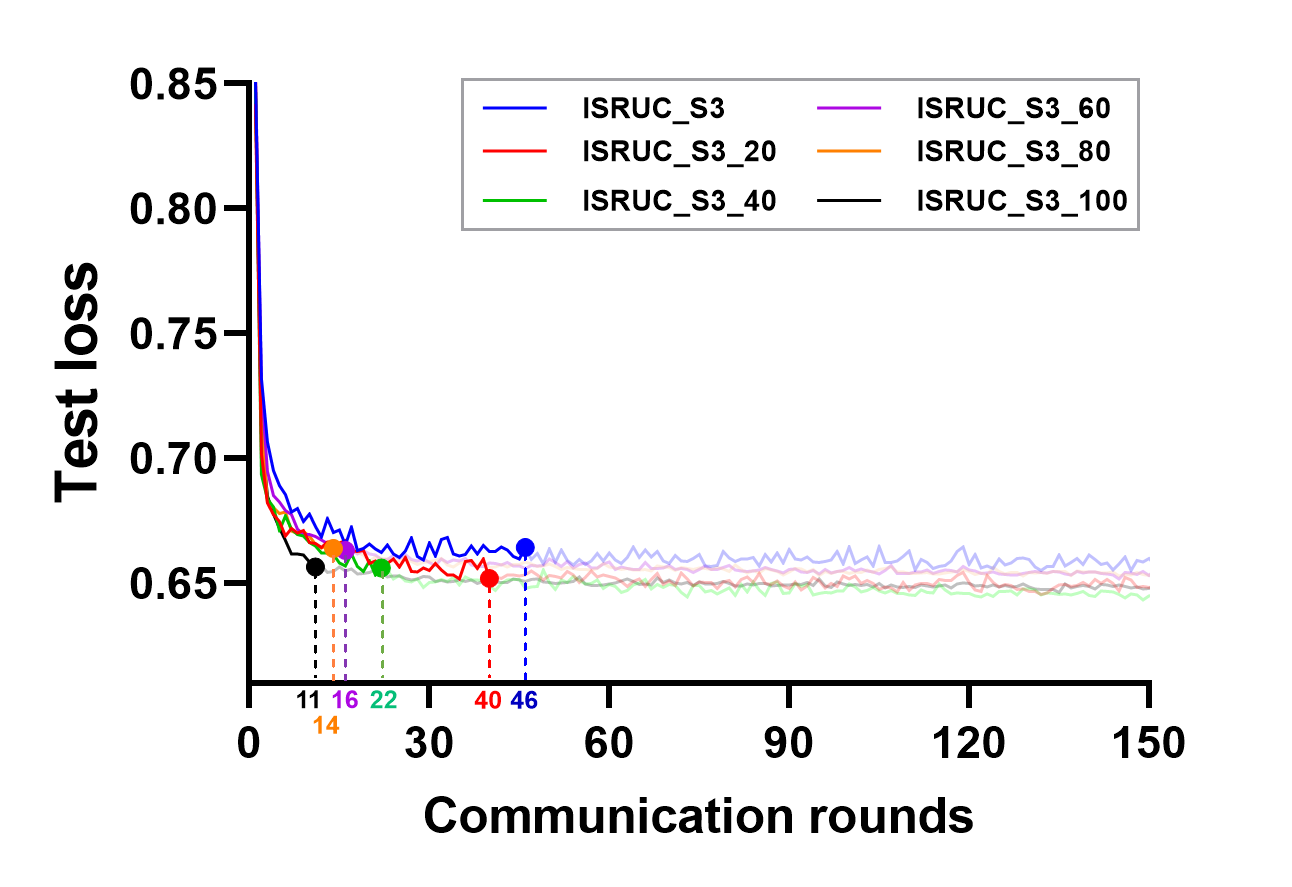} 
\caption{\revise{Convergence of loss on different scales of datasets.}}
\label{fig:diffscale}
\end{figure}
Fig.~\ref{fig:FLResult} reports the performance of the proposed framework and baseline methods in terms of the best F1 score and its changes over communication rounds. Fig.~\ref{fig:FLResult}a and Fig.~\ref{fig:FLResult}b show the F1 test score curve over communication rounds. It can be observed that FedAvg and FedP experience a slightly stronger oscillation on both datasets, while FedRel maintains a better performance throughout the training process. Compared with FedAtt on ISRUC\_S3, FedRel demonstrates a faster convergence speed while maintaining a higher score. When it comes to the smaller dataset SHL\_Small, it can be seen that the FedRel can converge even faster and demonstrate a more stable performance than other baselines (around 40 communication rounds for FedRel). In general, the numerical results in Fig.~\ref{fig:FLResult} demonstrate that, under the same model setup, FedRel achieves the best performance compared with baselines. 

Regarding the best F1 score metrics, FedRel is better than the Centralized by around 1.8\% while having around 3\% improvement compared with FedAvg on ISRUC\_S3 (Fig.~\ref{fig:FLResult}c). Compared with FedP and FedAtt, FedRel also demonstrates a clear increase. It shows similar results on SHL\_Small dataset (Fig.~\ref{fig:FLResult}d), whereas FedRel still achieves the best results compared with the other baselines and even enjoys a more considerable increase compared with FedP (around 3\%). To sum up, the promising performance of FedRel is credited to the design of distribution-driven relevance calculation, which helps facilitate the convergence speed by adaptively calibrating the data dispersion between each participant and the estimated global. The integrated attentive weight aggregations on the server side lead to the correct convergence direction and make the global model a more representative one. 

\begin{figure}[t]
\centering
\includegraphics[width=0.7\linewidth]{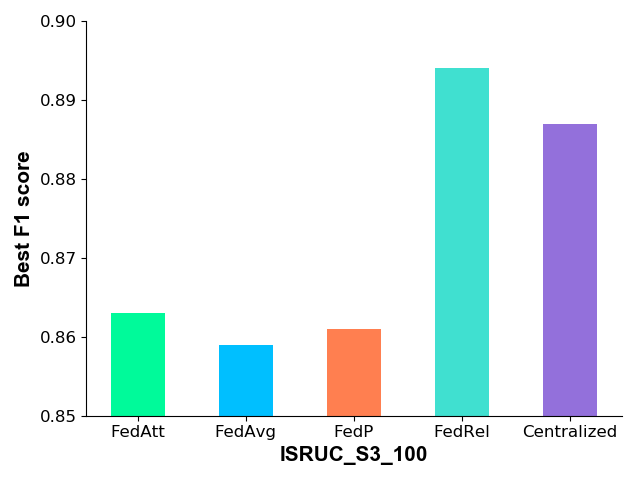} 
\caption{\revise{Comparison of F1 on ISRUC\_S3\_100.}}

\label{fig:isu_100}
\end{figure}

\revise{We also explored the convergence stability with the increase of dataset sizes. Specifically, FedRel is further tested on ISRUC\_S3, ISRUC\_S3\_20, ISRUC\_S3\_40, ISRUC\_S3\_60, ISRUC\_S3\_80 and ISRUC\_S3\_100, in which the ISRUC\_S3\_X indicates the data is collected from X subjects. As shown in Fig.~\ref{fig:diffscale}, we noticed a faster convergence speed when training samples increased. The reason is two-fold:1) each participant is able to utilise more local data to update the respective model at each communication round, thus speeding up the training process; 2) the increase of local data helps generate the high-quality latent data representation at each participant, which facilitates the relevance score of each participant adapts to the equilibrium state faster. The colored values along the x-axis in Fig.~\ref{fig:diffscale} refer to the convergence points of each dataset. Among them, ISRUC\_S3\_100 reaches the convergence point the fastest in 11 communication rounds, while ISRUC\_S3 only converges after 46 rounds of communication between participants and the server.
To verify the robustness of the FedRel on a much larger dataset, we further conduct the experiment on ISRUC\_S3\_100, which records the real-world polysomnography (PSG) data collected from 100 subjects and is ten times larger than ISRUC\_S3. Compared with smaller dataset ISRUC with marginal performance improvements (less than 2\%), the experiment results on ISRUC\_S3\_100 show that FedRel enlarges the performance gap over other FL baselines (shown in Fig.~\ref{fig:isu_100}), with a greater than 3.9\% performance gain compared with FedP, and an overall 3.4\% performance gain on average compared with other FL baselines.
This result demonstrates consistent robustness in the performance of our proposed framework when the dataset grows dramatically.}

\begin{table}[]
\caption{Performance comparison of different participants (best Accuracy (ACC) and F1 score at different communication rounds).}
\label{clients_num}
\centering

\begin{tabular}{|cl|rrrrr|}
\hline
\multicolumn{2}{|c|}{\multirow{2}{*}{}}                   & \multicolumn{5}{c|}{\# of Participant}                                                                                                          \\ \cline{3-7} 
\multicolumn{2}{|c|}{}                                    & \multicolumn{1}{c|}{2}      & \multicolumn{1}{c|}{3}      & \multicolumn{1}{c|}{5}      & \multicolumn{1}{c|}{10}     & \multicolumn{1}{c|}{50} \\ \hline
\multicolumn{1}{|c|}{\multirow{4}{*}{ISRUC\_S3}}  & ACC   & \multicolumn{1}{r|}{0.8736} & \multicolumn{1}{r|}{0.8722} & \multicolumn{1}{r|}{0.8698} & \multicolumn{1}{r|}{0.8664} & 0.8613                  \\ \cline{2-7} 
\multicolumn{1}{|c|}{}                            & Round & \multicolumn{1}{r|}{28}     & \multicolumn{1}{r|}{55}     & \multicolumn{1}{r|}{72}     & \multicolumn{1}{r|}{101}    & 167                     \\ \cline{2-7} 
\multicolumn{1}{|c|}{}                            & F1    & \multicolumn{1}{r|}{0.8558} & \multicolumn{1}{r|}{0.8547} & \multicolumn{1}{r|}{0.8431} & \multicolumn{1}{r|}{0.8423} & 0.8406                  \\ \cline{2-7} 
\multicolumn{1}{|c|}{}                            & Round & \multicolumn{1}{r|}{33}     & \multicolumn{1}{r|}{52}     & \multicolumn{1}{r|}{73}     & \multicolumn{1}{r|}{98}     & 161                     \\ \hline\hline
\multicolumn{1}{|l|}{\multirow{4}{*}{SHL\_Small}} & ACC   & \multicolumn{1}{r|}{0.9889} & \multicolumn{1}{r|}{0.9883} & \multicolumn{1}{r|}{0.9886} & \multicolumn{1}{r|}{0.9881} & 0.9824                  \\ \cline{2-7} 
\multicolumn{1}{|l|}{}                            & Round & \multicolumn{1}{r|}{47}     & \multicolumn{1}{r|}{72}     & \multicolumn{1}{r|}{128}    & \multicolumn{1}{r|}{141}    & 157                     \\ \cline{2-7} 
\multicolumn{1}{|l|}{}                            & F1    & \multicolumn{1}{r|}{0.9865} & \multicolumn{1}{r|}{0.9859} & \multicolumn{1}{r|}{0.9816} & \multicolumn{1}{r|}{0.9814} & 0.9778                  \\ \cline{2-7} 
\multicolumn{1}{|l|}{}                            & Round & \multicolumn{1}{r|}{44}     & \multicolumn{1}{r|}{77}     & \multicolumn{1}{r|}{128}    & \multicolumn{1}{r|}{135}    & 162                     \\ \hline
\end{tabular}

\end{table}

We further investigated the effect of the number of participants on the performance of FedRel. As shown in Table~\ref{clients_num}, we set the number of participants to be \{2, 3, 5, 10, 50\} in our framework and record data based on two metrics, i.e., performance (represented by the best Accuracy and F1 Scores) and communication rounds. Note that the training data being assigned to each participant follows the partial non-IID data distribution. It can be clearly observed that the performance of the framework decreases slightly with the increase in the number of participants on both datasets, and the degradation is acceptable (around 1\% in ISRUC\_S3 and negligible in SHL\_Small). Such experimental results are attributed to severe non-IID skewness caused by the discrepancy in the data distributions among participants. In other words, the growing number of participants inevitably leads to the higher possibility of difference and randomness in data partitions, which in turn develops the weight divergence phenomenon during training and requires more communication rounds for the model to gain the best performance~\cite{zhao2018federated}. Overall, the results in Table~\ref{clients_num} demonstrate the promising scalability of our proposed framework and its potential to be deployed in a real-world environment. \newline
\textbf{DIIG Performance (RQ2):} It is worth mentioning that the comparison between DIIG and other static correlation functions is intractable as DIIG considers the concept of inter-intra and takes at least two temporal graph snippets as input. Static correlation functions can only draw the connectivity of nodes on each spatial graph at one time step. As depicted in Table~\ref{tab:ablation_table}, the DIIG, regardless of combined feature transformation nets and GNNs, and the selections of time window $w$, achieves performance improvements on graph-level classification tasks compared with most static correlation methods. Intuitively, this result verifies that the inter-dependencies in the temporal dimension contain valuable topological information from the GNN's perspective, which is critical and helpful for learning the respective embedding of nodes and graphs. 
\subsubsection{Comprehensive Ablation Study}
We conduct comprehensive ablation studies in this part, while taking a close look at each module in our framework. \newline
\textbf{Spatial-temporal Data Processing (RQ3):}
As shown in Fig~\ref{fig:process}, the performance of different models adopted to generate initial node features is first studied. We implement different feature transformation models whiling having other modules fixed for fair comparison. To be more specifically, the results of CNN\_GNN\_FedRel, MLP-Mixer\_GNN\_FedRel, ConvLSTM\_GNN\_FedRel, Bi-LSTM\_GNN\_FedRel and Transformer\_GNN\_FedRel are gathered for comparison.

It can be found in Table~\ref{tab:ablation_table} that the initial node features generated by CNN helps our framework achieve the best results in ISRUC\_S3 dataset. Specifically, CNN improves by 12.39\% on average across different GNN models. On the other hand, MLP-Mixer generates a better quality of initial node features than CNN in the SHL\_Small dataset, achieving an average 9.8\% improvement. It is found that for smaller spatial-temporal dataset (SHL\_Small with hours of temporal sequence with only four labels), the MLP-Mixer is able to extract the critical features much more effective compared with other complex models.\newline
\textbf{GNN Models for Graph Learning (RQ4):}
When it comes to discovering the best match of the GNN model in DIIG, we first fix the choices of feature transformation net for each dataset and examine the results when deploying different GNN models. It can be observed that, in most cases, GAT achieves the best results compared with other GNN models. It proves that the attention-based weighted message passing in GAT also excels in the graphs containing only a small number of nodes (10 nodes per graph in ISRUC\_S3 and four nodes per graph in SHL\_Small).\newline
\textbf{Impact of Time Window $w$ (RQ5):} To answer \textbf{RQ5}, the effect of time window $w$ in DIIG is analysed. As observed in Table~\ref{tab:ablation_table}, the DIIG generally achieves the best results when $w=2$ or $w=3$, and it varies slightly based on the combinations of feature transformation nets and GNN models. We take a close look at the CNN\_GAT and MLP-Mixer\_GAT combinations, both of which achieve the best results in their respective datasets. For CNN\_GAT in ISRUC\_S3, DIIG reaches the best at $w=3$, while MLP-Mixer\_GAT, on the other hand, attains the most prominent result when $w=2$. This result attributes to the length of time sequence in each dataset. In other words, ISRUC\_S3 collects days of spatial-temporal signal sequences, it is thus less sensitive to the slight increase of the $w$ (from 2 to 3). SHL\_Small, on the other hand, composes only a fraction of the ISRUC\_S3's sequence length and is much easier to cause the overfitting problem with the increase of $w$. Therefore, we use CNN\_GAT\_DIIG($w=3$) and MLP-Mixer\_GAT\_DIIG($w=2$) as the ideal setup of each dataset in our experiment.

\section{Conclusion and Future Work}
In this paper, we propose an adaptive federated relevance framework for collaborative spatial-temporal graph learning. The feature transformation net module first extracts the important spatial-temporal feature information from the raw input data, followed by a dynamic inter-intral graph (DIIG) module to generate the spatial-temporal graphs while capturing inter-dependencies and dynamic temporal changes across these graphs. Most importantly, the federated relevance module facilitates the collaborative training of DIIGs from different participants, enabling an attentive weight aggregation on the basis of diverse data distributions. The effectiveness and superiority of the proposed framework are verified through extensive experiments with a variety of variants and baseline methods. Detailed ablation studies also demonstrate the design rationale of all key components. Finally, we believe that our proposed approach provides a general framework for better exploiting the spatial-temporal data in graph forms from different participants while respecting data privacy at the same time. In future work, we will explore new strategies that could take advantage of both feature and label space when it comes to calculating the local data distribution representations. 

\ifCLASSOPTIONcaptionsoff
  \newpage
\fi



%

\bibliographystyle{IEEEtran}
\bibliography{IEEEabrv,ref}











\newpage

\vfill

\end{document}